\documentclass[onecolumn]{article}
\usepackage{spconf,amsmath,graphicx, amsfonts}
\usepackage{xurl,mathrsfs}
\usepackage[hidelinks]{hyperref}
\usepackage[ruled,vlined,linesnumbered]{algorithm2e}
\usepackage{fancyhdr}
\usepackage{soul,color}

\title{Policy Augmentation: An Exploration Strategy for Faster Convergence of Deep Reinforcement Learning Algorithms}
\name{Arash Mahyari}
\address{Florida Institute for Human and Machine Cognition (IHMC) \\ 40 South Alcaniz St. Pensacola, FL 32502 \\ \textit{email}: amahyari@ihmc.org }
\begin{document}
%\ninept
%

\setlength{\textfloatsep}{10pt plus 0.0pt minus 0.0pt}
\maketitle
\begin{abstract}
Despite advancements in deep reinforcement learning algorithms, developing an effective exploration strategy is still an open problem. Most existing exploration strategies either are based on simple heuristics, or require the model of the environment, or train additional deep neural networks to generate imagination-augmented paths. In this paper, a revolutionary algorithm, called \textit{Policy Augmentation}, is introduced. \textit{Policy Augmentation} is based on a newly developed inductive matrix completion method. The proposed algorithm augments the values of unexplored state-action pairs, helping the agent take actions that will result in high-value returns while the agent is in the early episodes. Training deep reinforcement learning algorithms with high-value rollouts leads to the faster convergence of deep reinforcement learning algorithms. Our experiments show 
the superior performance of \textit{Policy Augmentation}. The code can be found at: \url{https://github.com/arashmahyari/PolicyAugmentation}.
\end{abstract}
\section{Introduction}
\label{sec:intro}

Reinforcement learning (RL) has shown significant success in recent years \cite{mnih2013playing, mnih2015human, mnih2016asynchronous,  fujimoto2019off, silver2017mastering}. RL studies how an agent can maximize its cumulative rewards in a previously unknown environment through experience. RL algorithms maximize their long-term cumulative rewards by exploring new states, and their short-term (immediate) cumulative rewards by exploiting their learned policy. 

The main challenge with RL algorithms is balancing between exploration and exploitation. While exploitation is critical to maximize the agent's immediate reward, the exploration ensures that the agent's overall (long-term) reward is maximized. Agents must explore most states to learn which states are more rewarding. However, the random exploration of the environment leads to a long convergence time. Depending on the problem and the environment, RL algorithms may require millions of episodes to explore all states and learn the policy. The $\epsilon$-greedy algorithm--a simple exploration strategy--takes either random actions (exploration) or actions according to the policy (exploitation). The simple strategies such as $\epsilon$-greedy well behave in situations where the reward is well-shaped. In more complicated tasks with sparse rewards, these strategies are not efficient. 

\vspace{-0.5mm}

More recently, several exploration strategies, \textit{e.g.}, curiosity-driven exploration \cite{zhelo2018curiosity}, count-based exploration \cite{tang2017exploration}, etc., have been proposed to improve the exploration of RL algorithms. The Uniform sampling strategy draws a random sample uniformly from the stored samples to update the action-value function--$Q$ functions \cite{mnih2015human}. Trust region policy optimization (TRPO) was proposed to generate a number of trajectories and randomly select a subset of states from these trajectories \cite{schulman2015trust}. The count-based exploration strategy  \cite{tang2017exploration} encourages agents to explore the less-visited states. 
These algorithms fail to perform optimally for tasks with sparse rewards. Besides, they simulate the future returns of rollout sets according to the current approximated policy. The performance of these algorithms depends on the approximated action-value functions according to previously explored state-action pairs. While these strategies are different from each other, they share a common idea that agents must be encouraged to visit less-explored states. These strategies take actions randomly from less-visited states, which may or may not lead to a reward. The existing exploration strategies do not encourage exploring state-action pairs that will return higher values.

\vspace{-0.5mm}

% These strategies encourage exploring all state-action pairs. They prioritize state-action pairs that have been less visited. The reason behind these strategies is to increase the number of samples from those states. These strategies take actions randomly from less-visited states. For example, if the agent has explored one state in a two-dimensional maze and has not explored three of its neighboring states, the agent will choose one of the three unexplored states randomly (in curiosity-driven, count-based, etc. methods) for exploration. The taken action may or may not lead to a reward. Thus, the exploration is still random in some sense. These strategies do not encourage exploring state-action pairs that will return higher values, resulting in a longer convergence time. 

%\vspace{-2mm}

\begin{figure}[t]
    \centering
    \includegraphics[width=5.2cm]{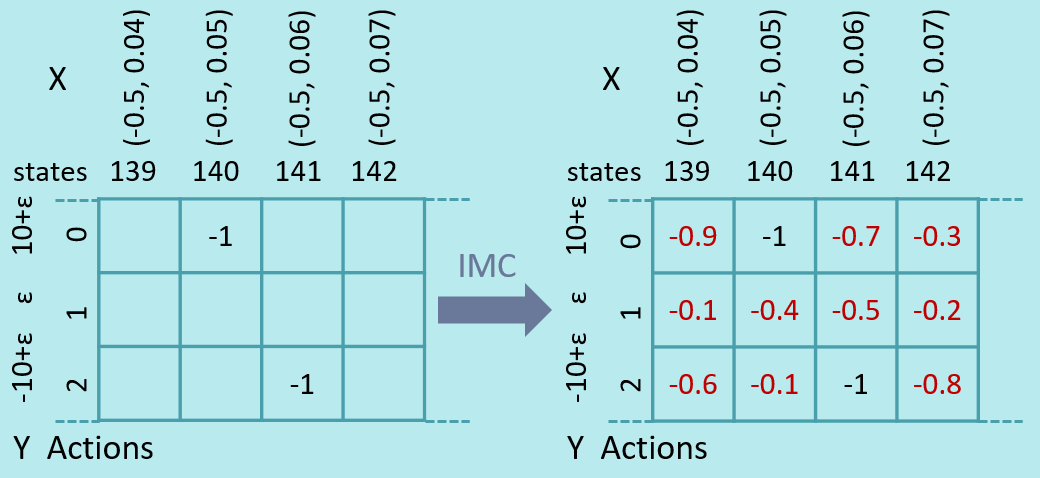}
    \vspace{-4mm}
    \caption{Augmenting $Q$ of \textit{Mountain Car} with IMC.}
    \label{fig:IMC}
\end{figure}

In this paper, we propose \textit{Policy Augmentation} to predict the values of unexplored state-action pairs by augmenting the action-value function, $Q$, using Inductive Matrix Completion (IMC). Matrix completion recovers the underlying low-rank matrix from a given subset of its entities \cite{dai2011subspace}. IMC extends this idea to the association inference prediction using the available side information about the columns and rows of the underlying low-dimensional matrix \cite{biswas2019robust}. In this paper, IMC recovers the $Q$ function from the limited observed entities--the limited number of explored state-action pairs--using the features of states and actions. Figure~\ref{fig:IMC} shows an example of recovered $Q$ function for the \textit{Mountain car} environment. The agent, then, takes actions according to the recovered action-value functions--the augmented policy. Taking actions according to the recovered $Q$ function results in generating high-value rollouts for training deep reinforcement learning (DRL) algorithms. The purpose of \textit{Policy Augmentation} is to generate high-value rollouts and train the agent with them during early episodes, which reduces the training time significantly. After training with high-value rollouts, the agent explores and exploits according to the existing DRL approaches, \textit{e.g.}, Proximal Policy Optimization (PPO) \cite{schulman2017proximal}.

% Once high-value rollouts are collected, the agent explores and exploits according to the existing DRL approaches, \textit{e.g.}, Proximal Policy Optimization (PPO) \cite{schulman2017proximal}. 
%\vspace{-5mm}

%\vspace{-2mm}

Fig.~\ref{fig:mountaincar}.a shows the classical \textit{Mountain Car} control problem \cite{moore1990efficient}. The environment is described in details in Section~\ref{sec:results}. The optimal policy must accelerate to the right when the car moves in the direction of the positive velocity, and accelerate to the left when the car is moving in the direction of the negative velocity. The existing exploration methods may apply negative, positive, or zero force when the car moves in the direction of the positive velocity. \textit{Policy Augmentation} explores the pairs that are predicted to have high values, \textit{e.g.}, positive force in the direction of the positive velocity. Our proposed strategy is different from \cite{racaniere2017imagination}, in that our method requires neither training nor the model of the environment, whereas the deep learning model used in \cite{racaniere2017imagination} requires  training.

% . Our method is able to augment the policy from the first episode, while the Imagination-augmented  agent\cite{racaniere2017imagination} requires several episodes until the model is trained. Moreover, the Imagination-augmented  agent\cite{racaniere2017imagination} assumes a learned environment model.

% The $Q$ function changes with time, and some state-action pairs that would be thought to be high value may become low value and vise versa. The goal is to prioritize the exploration of the pairs that we think would return higher values based on their side information (position, velocity, force) and the observed rewards so far. 

% \begin{figure}[t]
%     \centering
%     \includegraphics[width=4cm]{figures/mountaincar.jpg}
%     \caption{The classical \textit{mountain car} control problem \cite{moore1990efficient}. Action takes on any value between $-1$ and $+1$, the position of the car is continuous with value in $(-1.2, 0.6)$, and the velocity of the car is a continuous value in $(-0.07, 0.07)$. }
%     \label{fig:mountaincar}
% \end{figure}

\vspace{-2mm}

\section{Reinforcement Learning}

\vspace{-1mm}

 An infinite-horizon discounted Markov decision process (MDP) is defined by  $(\mathscr{S}; \mathscr{A}; \mathscr{P}, r, \rho_0, \gamma)$, where $\mathscr{S}$ is a set of finite states, $\mathscr{A}$ is a finite set of actions, $\mathscr{P}: \mathscr{S} \times \mathscr{A} \times \mathscr{S} \rightarrow \mathbb{R}$ is the transition probability distribution, $r: \mathscr{S} \rightarrow \mathbb{R}$ is the reward function, $\rho_0: \mathscr{S} \rightarrow \mathbb{R}$ is the distribution of the initial state ${s}_0$, and $\gamma \in (0,1)$ is the discount factor.

Let $\pi$ represents the stochastic policy of an agent that maps the action $a_t$ for a given $s_t$ to the future state $s_{t+1}$, and $R(\pi)$ represents the expected discounted reward of the policy: $ R(\pi)=\mathbb{E}_{s_0,a_0,\ldots} \left [ \sum_{t=0}^{\infty}{\gamma^t}r(s_t) \right ],$ where $s_0 \sim \rho_0(s_0)$, $a_t \sim \pi(a_t|s_t)$, and $s_{t+1} \sim \mathscr{P}(s_{t+1}|s_t, a_t)$. The state-action value function, $Q_\pi$, is defined as \cite{sutton2018reinforcement}: $Q_\pi(s_t,a_t)= \mathbb{E}_{s_{t+1},a_{t+1},\ldots} \left [ \sum_{l=0}^{\infty}{\gamma^l}r(s_{t+l}) \right ],$ where $a_t \sim \pi(a_t|s_t)$, and $s_{t+1} \sim \mathscr{P}(s_{t+1}|s_t,  a_t)$.

% \vspace{-1.5mm}
% \begin{equation*}
%     R(\pi)=\mathbb{E}_{s_0,a_0,\ldots} \left [ \sum_{t=0}^{\infty}{\gamma^t}r(s_t) \right ],
% \end{equation*}
% %$$ R(\pi)=\mathbb{E}_{s_0,a_0,\ldots} \left [ \sum_{t=0}^{\infty}{\gamma^t}r(s_t) \right ],$$
% \vspace{-1.5mm}

% \noindent where $s_0 \sim \rho_0(s_0)$, $a_t \sim \pi(a_t|s_t)$, and $s_{t+1} \sim \mathscr{P}(s_{t+1}|s_t, a_t)$. The state-action value function, $Q_\pi$, is defined as \cite{sutton2018reinforcement}: 
% \vspace{-1.5mm}
% \begin{equation*}
%     Q_\pi(s_t,a_t)= \mathbb{E}_{s_{t+1},a_{t+1},\ldots} \left [ \sum_{l=0}^{\infty}{\gamma^l}r(s_{t+l}) \right ],
% \end{equation*}
% \vspace{-1.5mm}
% \noindent where $a_t \sim \pi(a_t|s_t)$, and $s_{t+1} \sim \mathscr{P}(s_{t+1}|s_t,  a_t)$.

%$$ Q_\pi(s_t,a_t)= \mathbb{E}_{s_{t+1},a_{t+1},\ldots} \left [ \sum_{l=0}^{\infty}{\gamma^l}r(s_{t+l}) \right ],$$ where $a_t \sim \pi(a_t|s_t)$, and $s_{t+1} \sim \mathscr{P}(s_{t+1}|s_t,  a_t)$.

% \begin{equation*}
%     R(\pi)=\mathbb{E}_{s_0,a_0,\ldots} \left [ \sum_{t=0}^{\infty}{\gamma^t}r(s_t) \right ],
% \end{equation*}

%and the value function, $V_\pi$, are defined as:

% \begin{equation}
%     Q_\pi(s_t,a_t)= \mathbb{E}_{s_{t+1},a_{t+1},\ldots} \left [ \sum_{l=0}^{\infty}{\gamma^l}r(s_{t+l}) \right ], 
% \end{equation}

% \begin{equation}
%     V_\pi(s_t)= \mathbb{E}_{a_{t},s_{t+1},\ldots} \left [ \sum_{l=0}^{\infty}{\gamma^l}r(s_{t+l}) \right ], 
% \end{equation}

% \noindent where $a_t \sim \pi(a_t|s_t)$, and $s_{t+1} \sim \mathscr{P}(s_{t+1}|s_t,  a_t)$.

An agent interacts with an environment by sequentially selecting actions, observing states, and receiving rewards to find the optimal policy that maximizes the cumulative discounted reward $R_t=\sum_{l=0}^\infty{\gamma^l r_{t+l}}$. Different algorithms have been proposed to approximate $Q_\pi(s_t,a_t)$, \textit{e.g.}, $Q$-Learning, Deep $Q$-Learning \cite{sutton2018reinforcement, mnih2013playing}:

\vspace{-6mm}

\begin{multline}
    Q(s_{t},a_{t}) \leftarrow Q(s_t,a_t) + \alpha [ R_{t+1} \\ 
    + \gamma \max_a Q(s_{t+1},a)-Q(s_t,a_t) ] .
    \label{eq:qlearning}
\end{multline}
% Q(s_{t},a_{t}) \leftarrow Q(s_t,a_t) + \alpha \left [ R_{t+1} + \gamma \max_a Q(s_{t+1},a)-Q(s_t,a_t) \right ].
% \label{eq:qlearning}
% \end{equation}

% Q-learning is an off-policy RL method that approximates the Q function independent of the policy being followed \cite{sutton2018reinforcement}. The off-policy methods simplify the analysis and enable early convergence. The effect of policy is that it determines which state-action pairs are visited and updated and require to update all pairs for convergence \cite{sutton2018reinforcement}.

%\vspace{-5mm}

\section{Proposed Method}
\label{sec:proposed}

\vspace{-2mm}

In this section, we describe the proposed exploration strategy with \textit{Policy Augmentation}. The goal of the proposed method is to augment the policy based on what the agent has experienced so far using IMC methods. Our assumption is that the agent has neither any previous experiences, nor the model of the environment. We describes the proposed IMC method for recovering the $Q$ function in Section~\ref{sec:policy}. 

The goal of IMC methods is to recover the underlying low-rank matrix representation of the action-value function from the values of a small given subset of state-action pairs. If either state spaces or action spaces are continuous, we use hashing, \textit{e.g.}, SimHash function \cite{tang2017exploration}, to discretize their spaces and form the $Q$ matrix. 

Let $Q\in \mathbb{R}^{M \times N}$ be the actual action-value matrix that the agent will learn, and ${\hat Q} \in \mathbb{R}^{M \times N}$ represents the recovered action-value function by the IMC method, where $M$ is the total number of states and $N$ is the total number of actions. Let $\Omega_t$ represents the set of the observed entities of the action-value matrix--$Q_{\Omega_t}$--at time step $t$, and $Q_{\Omega_t}(s,a)$ refers to the value of the state $s$ and action $a$. After each time step, the agent receives a reward from the environment, $R_{t+1}$, that changes the value of either an observed entity of $Q_t$ or add a new member to the observed set $\Omega_t$.

% Let $Q_t$ be the action-value matrix at the time step $t$, and $Q_t(s,a)$ refers to the value of the state $s$ and action $a$ at time $t$. Let $\Omega_t$ represents the set of the observed entities of the action-value matrix ($Q_{\Omega_t}$) at time step $t$. After each time step, the agent receives a reward from the environment, $R_{t+1}$, that changes the value of either an observed entity of $Q_t$ or add a new member to the observed set $\Omega_t$. 

IMC methods incorporate additional information about actions and states, making them appropriate for predicting the values of state-action pairs that have not been explored. Let $X \in \mathbb{R}^{M \times m}$ and $Y \in \mathbb{R}^{N \times n}$ represent the state and action features (additional information), respectively. The $s$th row of $X$ represents the features of the $s$th state, and the $a$th row of $Y$ is the features of the $a$th action. For example, the position and the velocity of the vehicle in the \textit{Mountain Car} environment (Section~\ref{sec:results}) are hashed to $\{-1.2, -1.1, \ldots, 0.5, 0.6 \}$ and $\{-0.07, -0.06, \ldots$, $0.05, 0.06, 0.07\}$, respectively. The features of the states are $X=\{(1.2,-0.07), (1.2, -0.06), \ldots, (0.06, 0.6), (0.07,0.6) \}$. The features of the actions are $Y=\{-1,0,1\}$. These features are the basic knowledge of the environment.  

The state-action values (the entities of $Q_{\Omega_t}$) are unbounded and increase to large values, especially the states and the actions close to the initial state $s_0$. The large values of some entities of $Q_{\Omega_t}$ affect the performance of IMC methods. Thus, we normalize $Q_{\Omega_t}$ with the number of times the state-action pair is explored, $\tilde{Q}_t(s,a)=Q_t(s,a)/Nq_t(s,a)$, where $Nq_t(s,a)$ is the number of times $(s,a)$ is visited. 

% The advantage of using IMC methods is that they do not require training data (except for a few interactions). Thus, they can be integrated into all deep RL algorithms in new environments. 

%The action-value function, is a sparse matrix that gets filled in after each interaction. 

% IMC methods look at the interactions between states and actions in $Q$, and have been successfully deployed in recommendation systems to predict the user interests in given movies (or music) based on their interests in other movies (or music). The goal of IMC methods is to recover the underlying low-rank matrix representation of $Q \in \mathbb{R}^{M \times N}$ from the values of a small given subset of state-action pairs, where $M$ is the total number of states and $N$ is the total number of actions. Thus, each row corresponds to an action that agent can take and each column is associated with a state. IMC methods incorporate additional side information about actions and states, making them appropriate for predicting the values of state-action pairs for the states that have not been explored. Moreover, the side information about actions and states improve the accuracy of the estimated values.

Algorithm~\ref{alg:exploration} shows the proposed exploration strategy for training DRL algorithms. The strategy gets the DRL algorithm, state and action features ($X$ and $Y$), the total number of time steps ($\tau$), and initial state $s_0$ as its input. Because the goal is to generate high-value rollouts during the early episodes, we set a threshold $\tau_e$. The proposed algorithm uses the recovered augmented $\hat Q$ to take actions while $t<\tau_e$. During this time period, DRL collects high-value rollouts for training. For $t>\tau_e$, the exploration and exploitation of RL agents is similar to other previously proposed methods, \textit{e.g.}, PPO, TROPO, Curiosity-based, etc. Note that \textit{Policy Augmentation}  cannot replace a learned policy by DRL because it is only a prediction. It is used during the first $\tau_e$ episodes. 

The main assumption of matrix completion and IMC methods is that the underlying matrix--$Q$ function--is low rank. Thus, we calculate the IMC method for the first $\tau_q$ steps ($\tau_q<\tau_e$) while the $Q$ function is still low-rank. As the agent visits more state-action pairs, the rank of the $Q$ increases. The computational complexity of the proposed method is associated with the complexity of the proposed IMC method.

\begin{algorithm}[t]\label{alg:exploration}
\SetAlCapSkip{0ex}
\SetAlCapHSkip{0ex}
\caption{Exploration Strategy with Policy Augmentation}
\SetAlgoLined
\textbf{Input: }$X$, $Y$, RL Algorithm, $\tau$, $\tau_e$, $\tau_a$\\
{\textbf{Output: }}{Trained DRL algorithm} \\
\textbf{Initialization:} ${\tilde Q}_0=0$, $Nq=0$, $t=0$, $s_0$\\
%\KwResult{Write here the result }
\While{$t < \tau $}{
 \uIf{$t < \tau_e $}{$a_t=\arg \max {\hat Q}(s_t,:)$}
 \Else{$a_t$ from the DRL algorithm, \textit{e.g.}, PPO.}
 Take $a_t$, receive $s_{t+1}$ and $r_t$.\\
 Collect rollouts for training the DRL algorithm.\\ % off-policy or on-policy algorithms.\\
 \uIf{$t < \tau_q$}{
 Update ${\tilde Q}_t$ using Eq.~\ref{eq:qlearning} and $\tilde{Q}_t(s,a)=Q_t(s,a)/Nq_t(s,a)$.\\
  Initialize $U_t$ and $V_t$.\\
 \Repeat{convergence criterion is met}{
 Update $U_t$ using Eq.\ref{eq:solu1}. \\
 Update $V_t$ using Eq.\ref{eq:solu2}.} 
 Augment the policy ${\hat Q}=X^TU_tV_t^TY^T$.}
 Train the DRL algorithm using the rollouts. 
 }
\end{algorithm}

% \begin{algorithm}[t]\label{alg:exploration}
% \SetAlCapSkip{0ex}
% \SetAlCapHSkip{0ex}
% \caption{Exploration Strategy with Policy Augmentation}
% \SetAlgoLined
% \algorithmicindent\textbf{Input:}{$X$, $Y$, RL Algorithm, $\tau$, $\tau_e$, $\tau_a$} \\

% \algorithmicindent\textbf{Initialization:} ${\tilde Q}_0=0$, $Nq=0$, $t=0$, $s_0$\\
% %\KwResult{Write here the result }
% \While{$t < \tau $}{
%  \uIf{$t < \tau_e $}{$a_t=\arg \max {\hat Q}(s_t,:)$}
%  \Else{$a_t$ from the DRL algorithm, \textit{e.g.}, PPO.}\\
%  Take $a_t$, receive $s_{t+1}$ and $r_t$.\\
%  Collect rollouts for training the DRL algorithm.\\ % off-policy or on-policy algorithms.\\
%  Update ${\tilde Q}_t$ using Eq.~\ref{eq:qlearning} and $\tilde{Q}_t(s,a)=Q_t(s,a)/Nq_t(s,a)$.\\
%  \eIf{$t$ is odd}{Update $U_t$ using Eq.\ref{eq:solu1}}{Update $V_t$ using Eq.\ref{eq:solu2}}\\
%  \lIf{$t < \tau_q$}{Augment the policy ${\hat Q}=X^TU_tV_t^TY^T$.}\\
%  Train the DRL algorithm using the rollouts.} 
%  \algorithmicindent\textbf{Output:}{Trained DRL algorithm}
% \end{algorithm}

\vspace{-2mm}

\subsection{Policy Augmentation}
\label{sec:policy}

\vspace{-1mm}

The proposed IMC method recovers the $Q$ function from a given set of observed state-action pairs. We assume that $Q_{t}$ is a low-rank matrix (during the first $\tau_q$ time steps) while the agent is in the early stages of exploration and the set $\Omega_t$ is very small: $|\Omega_t| \ll M \times N$, where $|\Omega_t|$ is the size of the set $\Omega_t$. The normalized state-action value matrix is written as ${\tilde Q}_{t}=XW_{t}Y^T$, where $W_{t}$ is an unknown, low-rank matrix $W_{t}=U_{t} V_{t}^T \in \mathbb{R}^{m \times n}$ with the unknown rank $r\ll min(m,n)$. Our goal is to recover ${\hat Q}_{t}$ from ${\tilde Q}_{\Omega_t}$ by solving $\min_{U_t,V_t} \hspace{2mm} \| {\tilde Q}_{\Omega_{t}}-XU_t V_t^T Y^T\|_F^2=\min_{U_t,V_t} \hspace{1mm} tr(({\tilde Q}-XU_t V_t^T Y^T)^T ({\tilde Q}_{\Omega_{t}}-XU_t V_t^T Y^T))$ \cite{dai2011subspace}. 
There exists a unique solution to this optimization problem if at least $nm \log(nm)$ state-action values are explored \cite{jain2013provable}, and $U_t$ and $V_t$ are incoherent \cite{keshavan2010matrix} \footnote{Interested readers may refer to \cite{jain2013provable, keshavan2010matrix} for the proof.}.

%Because $X$ and $Y$ are known matrices, we impose the inchoherency condition on $U_t$ and $V_t$. 

The contribution of our proposed IMC method is that it racks the changing set $\Omega_t$ as new pairs are revealed during the exploration. Moreover, the values of the observed set $\Omega_t$ changes with time because the agent receives more rewards (either positive, negative, or neutral) from the environment for repeating the previously taken actions (Eq.~\ref{eq:qlearning}). Following the definition in \cite{narayanamurthy2019provable, dai2011subspace, hong2014geodesic}, we consider ${\tilde Q}_{1}, {\tilde Q}_{2}, \ldots, {\tilde Q}_{t}$ as a set of linear subspaces on the Grassmann manifold, forming a geodesic. ${\tilde Q}_{i}$ is a point in the Grassmann manifold and is represented as a pair $(U_i,V_i)$ \cite{keshavan2010matrix, dai2011subspace}. At each time step, only one element of ${Q}_{t}$ is changing. Thus, the linear subspaces of ${\tilde Q}_{t-1}$ and ${\tilde Q}_{t}$ remain close to each other. The distance between two subspaces is defined by the canonical distance (geodesic or arc-length distance) $d^2={d(U_t,U_{t-1})^2+d(V_t,V_{t-1})^2}$ \cite{keshavan2010matrix}. We use a set of principal angles $\{\theta_j\}_{j=1}^r$, $r=min(m,n)$ to quantify $d(U_t,U_{t-1})$ and $d(V_t,V_{t-1})$ as the projection distance: $d^2(U_t,U_{t-1})=\sum_{k=1}^r{\sin^2 \theta_k}=r-tr(U_tU_{t}^TU_{t-1}U_{t-1}^T)$, and $d^2(V_t,V_{t-1})=r-tr(V_tV_{t}^TV_{t-1}V_{t-1}^T)$ \cite{dong2013clustering}. The augmented action-value matrix, $\hat Q$ is recovered by minimizing the cost function $J$ with respect to $U_t$ and $V_t$ \cite{biswas2019robust}:

% \begin{equation}\label{eq:opt}
%  \begin{array}{cc}
%     J=\min_{U_t,S_t,V_t} \hspace{2mm} \| Q_{\Omega_t}-P_{\Omega_t}(XU_t S_t V_t^T Y^T)\|_F^2 \\ 
%     +d^2(U_t,U_{t-1})+d^2(V_t,V_{t-1})\\
%     =\min_{U_t,S_t,V_t} \hspace{2mm} \| Q_{\Omega_t}-P_{\Omega_t}(XU_t S_t V_t^T Y^T)\|_F^2 \\ 
%     +{1 \over 2} \|U_tU_{t}^T-U_{t-1}U_{t-1}^T\|_F^2\\
%     +{1 \over 2} \|V_tV_{t}^T-V_{t-1}V_{t-1}^T\|_F^2,
%      \end{array}
% \end{equation}

\vspace{-6mm}

\begin{multline}\label{eq:opt}
    J= tr(({\tilde Q}-XU_t V_t^T Y^T)^T ({\tilde Q}_{\Omega_{t}}-XU_t V_t^T Y^T)) \\ 
    +r-tr(U_tU_{t}^TU_{t-1}U_{t-1}^T)
    +r-tr(V_tV_{t}^TV_{t-1}V_{t-1}^T).
\end{multline}

% \noindent where $\|.\|_F$ is the Forbenius norm, and $\lambda$ is the regularization parameter. 

\noindent By taking the partial derivative of the cost function $J$ with respect to $U_t$ and $V_t$ and set them to zero \cite{dai2011subspace}, the closed form expressions are obtained. Then, the values of $U_t$ and $V_t$ are updated iteratively similar to \cite{biswas2019robust}. The matrices $U_t$ and $V_t$ are initialized with random dense matrices. The algorithms updates $U_t$ and $V_t$ in Eq.\ref{eq:solu1} and~\ref{eq:solu2} alternatively until the convergence criteria meets. 
\vspace{-3mm}
%\vspace{-3mm}

% \begin{equation}
%     \begin{array}{cc}
%          {\frac{\partial J}{\partial U_t}}=-2 P_{\Omega_t}({\tilde Q}_{\Omega_{t}}-XU_t V_t^T Y^T) XYV_t  
%     +2U_{t-1} U_{t-1}^T U_{t} \\
%          {\frac{\partial J}{\partial V_t}}= -2 P_{\Omega_t}({\tilde Q}_{\Omega_{t}}-XU_t V_t^T Y^T) X U_t Y^T +2V_{t-1} V_{t-1}^T V_{t} 
%     \end{array}
% \end{equation}

% {\small
% \begin{equation*}
%     \begin{array}{cc}
%          {\frac{\partial J}{\partial U_t}}=-2X^T{\tilde Q}YV_t+2X^TXU_tV_t^TY^TYV_t-2\lambda_1 U_{t-1} U_{t-1}^T U_{t} \\
      
%          {\frac{\partial J}{\partial V_t}}= -2 Y^T{\tilde Q}^TXU_t+2Y^TYV_tU_t^TX^TXU_t-2\lambda_2 V_{t-1} V_{t-1}^T V_{t} 
%     \end{array}
% \end{equation*}
% }

% \vspace{-4mm}

\begin{align}
& \resizebox{0.90 \hsize}{!}{ $ U_t{(i,j)}=U_t{(i,j)} \left [{\frac{X^T{\tilde Q}YV_t}{X^TXU_tV_t^TY^TYV_t-{\lambda}_1 U_{t-1} U_{t-1}^T U_{t}} } \right]_{(i,j)} $ \label{eq:solu1}} \\
& \resizebox{0.90 \hsize}{!}{ $V_t{(i,j)}=V_t{(i,j)}  \left [{\frac{Y^T{\tilde Q}^TXU_t}{Y^TYV_tU_t^TX^TXU_t-{\lambda}_2 V_{t-1} V_{t-1}^T V_{t}} } \right]_{(i,j)}, $ \label{eq:solu2}}
\end{align}         

\vspace{-1mm}

\noindent where $(i,j)$ indicates the element of the matrix in the $i$th row and the $j$th column. We selected $\lambda_1=\lambda_2=1$ here.

\vspace{-2mm}

\section{Results}
\label{sec:results}

\vspace{-2mm}

The OpenAI Gym toolkit \cite{brockman2016openai} and the Stable-baseline package \cite{stable-baselines} are used to evaluate the proposed algorithm. Two environments, \textit{Mountain Car} and \textit{CartPole}, are selected. \textit{Policy Augmentation} is integrated with Proximal Policy Optimization (PPO) \cite{schulman2017proximal} (\textit{PPO+ Policy Augmentation}), and Deep Q-Learning (DQN) \cite{mnih2013playing} (\textit{DQN+ Policy Augmentation}). The performances of \textit{PPO+Policy Augmentation} and \textit{DQN+Policy Augmentation} are compared with PPO \cite{schulman2017proximal}, DQN \cite{mnih2013playing}, PPO+ Counting \cite{tang2017exploration}, and AttA2C (Att+ Curiosity) \cite{reizinger2020attention} in both environments. The experiment is repeated $10$ times with $10$ random initialization, and each time for $2,500,000$ rollouts. The performances of different methods and exploration strategies are compared using their cumulative rewards. Because the purpose of \textit{Policy Augmentation} is to boost the performance of DRL algorithms during the early episodes, we present the results of only a few first episodes. 
\newline
\textbf{\textit{CartPole}:} The agent playing \textit{CartPole} receives an observation with four elements from the environment: \textit{Cart Position} $\in [-4.8, 4.8]$, \textit{Cart Velocity} $\in (-\infty, \infty)$, \textit{Pole Angle} $\in [-24^o, 24^o]$, and \textit{Pole Angular Velocity} $\in (-\infty, \infty)$. The agent can push the cart either to the left ($0$) or to the right ($1$), and receives one reward for every step. The game terminates if either $|$\textit{Pole Angle} $|> 12^o$, $|$\textit{Cart Position} $|> 2.4$, or episode length is greater than $200$. At the start of each episode, the position of the cart is randomly drawn from the uniform distribution $[-0.05, 0.05]$. \textit{Cart Position}, \textit{Cart Velocity}, \textit{Pole Angle}, \textit{Pole Angular Velocity} are used as the state features--columns of the matrix $X$, in \textit{Policy Augmentation}. We used the vector $Y=[-10, 10]^T$ as the action features; $10$ is selected arbitrary to increase the difference between left and right actions. The positive and negative signs are the side information indicating the opposite direction of two actions.

\begin{figure}[t]
\begin{minipage}[b]{0.48\linewidth}
  \centering
  \centerline{\includegraphics[width=3.6cm,height=2.9cm]{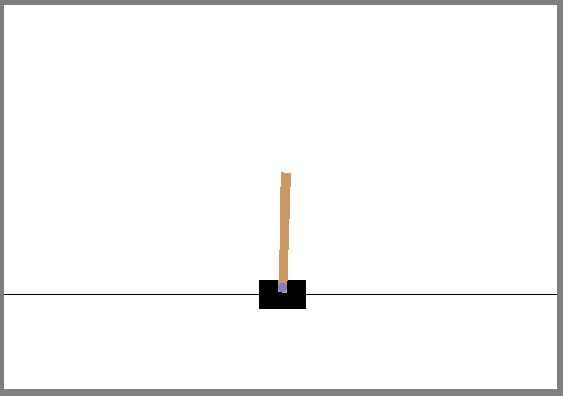}}
  \vspace{0.3cm}
  \centerline{(a)}%\medskip
\end{minipage}%
\hfill
\begin{minipage}[b]{0.52\linewidth}
  \centering
  \centerline{\includegraphics[width=4.5cm,height=3.3cm]{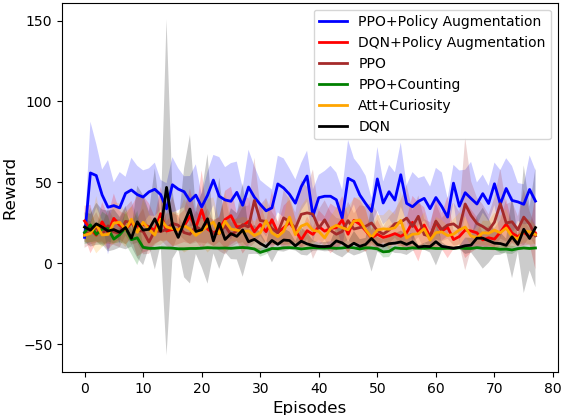}}
  %\vspace{0.1cm}
  \centerline{(b)}%\medskip
\end{minipage}
\vspace{-7mm}
\caption{(a) Cart Pole Environment; (b) cumulative rewards of each episode. The darker line is the mean of the cumulative rewards over $10$ repetitions. The shaded areas represent one standard deviation of the cumulative rewards.}
\label{fig:cartpole}

\end{figure}

Figure~\ref{fig:cartpole}.b shows the cumulative rewards of different agents playing the Cart Pole. The darker lines represent the mean of the cumulative rewards of $10$ repetitions, and the shaded areas represent one standard deviation of the cumulative rewards for $10$ repetitions. Overall, the \textit{PPO+ Policy Augmentation} has the best performance, and the \textit{DQN+Policy Augmentation} has a comparable performance with the PPO and the AttA2C (Att+ Curiosity) methods. PPO has a better performance than DQN. When PPO is combined with \textit{Policy Augmentation}, it shows superior performance.

\begin{figure}[t]
\begin{minipage}[b]{0.48\linewidth}
  \centering
  \centerline{\includegraphics[width=3.6cm,height=2.9cm]{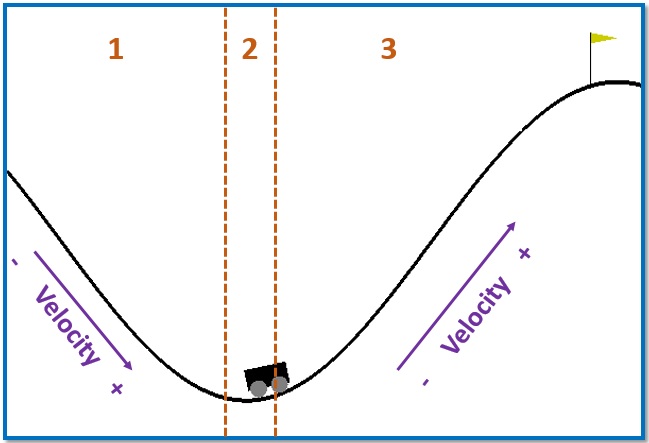}}
  %\centerline{\includegraphics[width=3.6cm,height=2.9cm]{figures/mountain_car.png}}
  \vspace{0.3cm}
  \centerline{(a)}%\medskip
\end{minipage}%
\hfill
\begin{minipage}[b]{0.52\linewidth}
  \centering
  \centerline{\includegraphics[width=4.5cm,height=3.3cm]{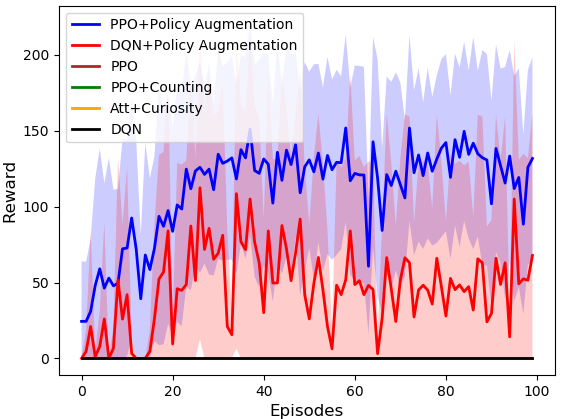}}
  %\vspace{0.1cm}
  \centerline{(b)}%\medskip
\end{minipage}
\vspace{-7mm}
\caption{(a) Mountain Car Environment; (b) cumulative rewards of each episode. The darker line is the mean of the cumulative rewards over $10$ repetitions. The shaded areas represent one standard deviation of the cumulative rewards.}
\label{fig:mountaincar}

\end{figure}

\textbf{\textit{MountainCar}:} The state of the \textit{Mountain Car} environment is two dimensional: \textit{Car Position} $\in [-1.2, 0.6]$ and \textit{Car Velocity} $\in (-0.07, 0.07)$. The agent can take either of three actions: Accelerate to the Left (0), Don't accelerate (1), and Accelerate to the Right (2). The negative force moves the car in the direction of the negative velocity in Fig.~\ref{fig:mountaincar}.a, and the positive force moves the car in the positive direction. At the start of each episode, the position of the cart is randomly drawn from the uniform distribution $[-0.6, -0.4]$. The agent receives $-1$ as its reward for every step it takes unless the car reaches the flag, when the agent receives the reward $10$. The game terminates if either $|$\textit{Car Position} $|> 0.5$, or episode length is greater than $200$. 

We used \textit{Car Position} and \textit{Car Velocity} as the columns of $X$ (state features), and $Y=[-10+\epsilon, 0+\epsilon, 10+\epsilon]^T$. $epsilon$ is a small non-zero value, \textit{e.g.}, $1$, that prevents multiplying $W$ by zero. 
Figure~\ref{fig:mountaincar}.b shows the cumulative rewards of different agents. The darker lines represent the mean of the cumulative rewards of $10$ repetitions, and the shaded areas represent one standard deviation of the cumulative rewards over $10$ repetitions. The \textit{PPO+ Policy Augmentation} and the \textit{DQN+Policy Augmentation} have the best and the second best performances, respectively. Other methods are not able to generate high-value rollouts during the first $100$ episodes.

The experiments showed that the performance of the proposed \textit{Policy Augmentation} exploration strategy depends on the first taken action. In the beginning, the $Q$ matrix is zero and the agent chooses randomly an action from all zero-valued actions. The estimation of $\hat Q$ values are affected by which action is taken in the first step. To achieve a superior performance, the proposed strategy observes the cumulative reward. If the cumulative rewards during the first few episodes are below a predefined value, $Q$ is reset to zero and the agent takes a new action randomly.

% We set up the \textit{Mountain Car} environment with sparse rewards. The agent does not receive any reward unless the car reaches the top of the mountain (the flag), when the agent receives the reward $10$. The agent receives a two-dimensional observation (state) from the environment: \textit{Car Position} $\in [-1.2, 0.6]$ and \textit{Car Velocity} $\in (-0.07, 0.07)$. The agent can take either of three actions at a time step: Accelerate to the Left (0), Don't accelerate (1), and Accelerate to the Right (2). At the start of each episode, the position of the cart is randomly drawn from the uniform distribution $[-0.6, -0.4]$. We gave an internal reward $-1$ for each step the agent 

%\vfill
%\pagebreak

\vspace{-1mm}

\section{Conclusion}
\label{sec:prior}
\vspace{-2mm}
\textit{Policy Augmentation} predicts the values of unexplored state-action pairs using IMC methods. By taking actions at early stages according to the predicted state-action values, agents collect rollouts with high values. Deep reinforcement learning algorithms trained with the high-value rollouts learn policies faster than existing exploration strategies. \textit{Policy Augmentation} results in faster training of deep reinforcement learning algorithms, which makes deploying deep reinforcement learning algorithms in new environments more practical. 

Future work will consider developing more computationally efficient IMC algorithms that update the previously augmented policy after each step from the new observed entity. The future work will also develop DRL algorithms that balance the number of high-value and low-value rollouts for training in environments with sparse rewards.

\vfill

%\newpage

% \section{REFERENCES}
% \label{sec:refs}

%\bibliographystyle{abbrv}
\bibliographystyle{IEEEbib}
\bibliography{strings,refs}

\end{document}